# Real-time Surgical Tools Recognition in Total Knee Arthroplasty Using Deep Neural Networks

Moazzem Hossain[1,3], Soichi Nishio[1], Takafumi Hiranaka[2] and Syoji Kobashi[1]

*University of Hyogo[1], Japan.*

*Takatsuki Hospital[2], Takatsuki*

*International University of Business Agriculture & Technology[3], Bangladesh*

*Abstract*—Total knee arthroplasty (TKA) is a commonly performed surgical procedure to mitigate knee pain and improve functions for people with knee arthritis. The procedure is complicated due to the different surgical tools used in the stages of surgery. The recognition of surgical tools in real-time can be a solution to simplify surgical procedures for the surgeon. Also, the presence and movement of tools in surgery are crucial information for the recognition of the operational phase and to identify the surgical workflow. Therefore, this research proposes the development of a real-time system for the recognition of surgical tools during surgery using a convolutional neural network (CNN). Surgeons wearing smart glasses can see essential information about tools during surgery that may reduce the complication of the procedures. To evaluate the performance of the proposed method, we calculated and compared the Mean Average Precision (MAP) with state-of-the-art methods which are fast R-CNN and deformable part models (DPM). We achieved 87.6% mAP which is better in comparison to the existing methods. With the additional improvements of our proposed method, it can be a future point of reference, also the baseline for operational phase recognition.

*Keywords*-Surgical tools recognition, TKA, Neural network, Faster R-CNN.

## I. INTRODUCTION

TKA is a safe and effective treatment to restore function and relieve pain in patients with arthritic knee[1]. As the population ages, there has been a trend towards an increasing number of patients with this pathological condition and a higher demand for TKA. Approximately 700,000 knee replacement procedures are performed every year only in the United States. The demand for TKA is assumed to grow by 673% to 3.48 million procedures by 2030 [2]. As the number increases, the search for new options that can help to improve TKA results and refine the surgical procedure is essential.

Computer-assisted orthopedic surgery (CAOS) is a field of technological evolution developed in particular in the last decade, and several surgical applications have emerged during this period. Computer-aided TKA is one of them that increase the surgeon's capabilities by presenting information during the surgery. In CAOS, the understanding of the surgical workflow and the extraction of the procedure in the operative phases is a potential step to reduce the surgical errors by giving instructions to the surgeon which steps to follow next. For surgical workflow recognition, real-time tools detection is the most crucial part to obtain information about surgery. The development of cognitive systems that automatically analyze the flow of surgical work would help surgeons and physicians, providing solutions for essential tasks in the operating room such as surgical phase recognition, evaluation, monitoring operation and optimizing the time of surgery.

One of the most significant challenges for image-based techniques for detecting surgical tools is the robustness and in particular wide range of surgical tools and conditions that may affect image quality and visibility of tools. With this paper, we studied the current state-of-the-art methods in image-based surgical tools detection and proposed a method that uses CNN focusing on the aspects of previous work. The extraction of information on the presence and movement of the instruments during surgery can be obtained through different detection methods like region based convolutional neural network (R-CNN), fast R-CNN [4], DPM, single shot multi-box detector (SSD) and region based fully connected network (R-FCN) on deep learning approach. We used faster R-CNN architecture to detect surgical tools in real-time and achieved better performance in comparison to state-of-the-art methods for tools detection.

Early work on detection, categorization, and monitoring of surgical instruments include those based on radiofrequency identification tags (RFIDs); segmentation, contour processing and 3D modeling; and the Viola-Jones detection framework [14]. Besides, deep learning approaches based on convolutional neural networks have demonstrated impressive performance in computer vision [15], and the included works [7, 8, 9, 22] take advantage of deep learning architectures to achieve state-of-the-art performance in surgical tools detection and phase recognition. As part of the M2CAI 2016 tools presence detection challenge [10], they have introduced a reference point for the detection of the presence of surgical tools. Although several existing studies refer to the detection of presence at the frame level, Sarikaya et al. perform the localization of surgical tools in robot-assisted surgical training videos, using multimodal convolutional neural networks [16]. The automated understanding of the surgical scene and the evaluation of skills are other areas of study. Some studies have involved specific components of surgical video analysis, including recognition of the operational phase and recognition of the activity. As part of the M2CAI 2016 Surgical Workflow Challenge, the works that include [17, 18, 19] are about the recognition of the operational phase in the cholecystectomy videos. Also, Zia et al. in particular, it analyzes the specific suture and knots videos by activity, using the typical, texture and frequency characteristics [20]. Similarly, Lalys et al. propose a framework using the hidden Markov model and the visual characteristics, such as shape, color, and texture, to identify operational tasks [14]. One limitation of these specific activity studies is that surgical training tasks differ substantially and do not accurately reflect operational performance in real surgical interventions. Our work builds on these prior contributions and uses region-based convolutional neural networks to detect the spatial bounds of tools, enabling more affluent and comprehensive assessment of surgical quality in real-world surgery.

This paper proposes a surgical tools detection method that exploits region based convolutional neural networks to perform the detection of surgical tools in real-time and significantly overcome previous work on the detection of surgical tools. This paper organized in the following sections. Sec. II introduces TKA focused in this study, and smart glass attached to surgeons. Sec. III proposes a tools detection method for TKA based on faster R-CNN from the smart glass video images. Sec. IV demonstrates the experimental results and Sec. V summaries this study.

## II. PRELIMINARIES

### A. Total Knee Arthroplasty

Total knee arthroplasty, also known as total knee replacement, is one of the most commonly performed orthopedic procedures. A variety of pathological conditions that affect the knee can be treated with TKA, which leads to pain relief and restoration of function of the knee. Advances in medical technology, including accelerometer-based navigation, patient-specific instrumentation, and robot-assisted total knee arthroplasty, have been developed to improve the accuracy and precision of TKA. In the entire surgery 27 procedures that require many tools around 120 of different categories. It is challenging to identify tools manually by the surgeon during the surgery and also for the surgical assistant. Our system will help the surgeon to detect surgical tools in real-time using the smart glass.

### B. Smart Glass

Smart glasses are wearable glasses that add information alongside or to what the wearer sees. The feature of this device is that it is lighter (about 50 g) than the conventional headset camera, and it is possible to use bidirectional communication in real-time. It is an innovative device (shown in fig.1) that allows surgeons to keep their eyes in the field of operation during procedures carried out under fluoroscopic control [3].

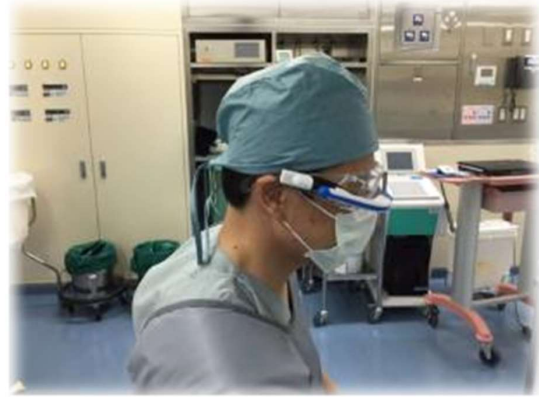

**Figure 1:** A surgeon wearing smart glass during TKA that can be used in emergency surgery and surgical training, including live transmission, as well as remote instructions and monitoring the conditions of surgery in real-time.

## III. PROPOSED METHOD

### A. Surgical tools recognition system

We proposed a surgical tools recognition system based on the architecture of Faster R-CNN [4], a region based convolutional neural network. We first convolve the input and get their convolutional feature maps that are calculated from TKA surgery video frames, and the output is the coordinates of bounding boxes around any of the surgical tools used in the surgery. We use the RGB image convolutional features, to train a Region Proposal Network (RPN) that generates object proposals. Region proposals are relative reference boxes centered at each sliding window. Each of these proposals has individual Objectness score. In our research, we proposed method shown in Fig. 2 based on the architecture of faster R-CNN.

The base network is a VGG-16 convolutional neural network with 16 convolutional layers, which extracts powerful visual features. At the top of this network, there is a region proposal network (RPN) that shares convolutional features with object detection networks. From each input image, the RPN generates region proposals that contain an object and features are pooled in these regions before moving on to a final classification and bounding box refinement network. The use of the RPN allows significant computational gains and detection accuracy concerning the previous related work, Fast R-CNN [4] and DPM [5].

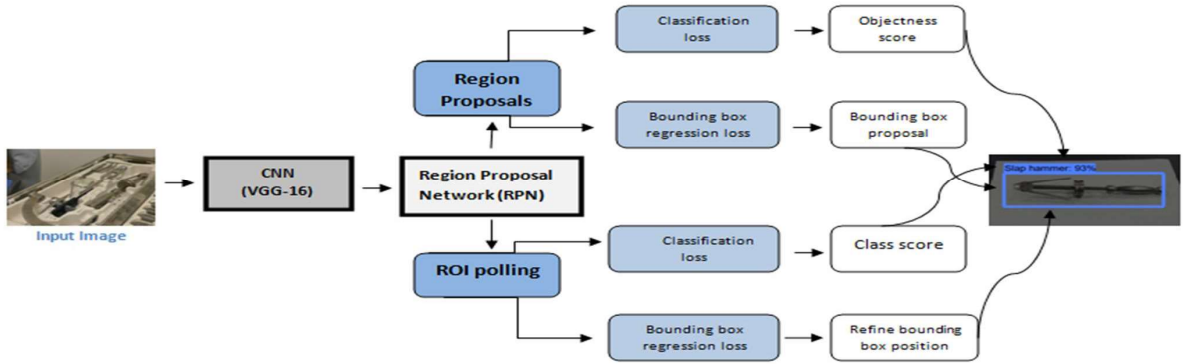

**Figure 3:** Our proposed method. The input of this network is the surgical video frame. The fastest R-CNN base network is a VGG-16 convolutional neuronal network. This is connected to a regional proposal network (RPN) that shares convolutional features with object detection networks. For each input image, the RPN generates region proposals that contain an object, and the characteristics are grouped in these regions before passing them to the final classification. The output is the bounding box positions of detected surgical tools.

The RPN is trained by optimizing the following loss function for each image:

$$L(\{p_i\}, \{t_i\}) = \frac{1}{N_{cls}} \sum_i L_{cls(p_i, p_i^*)} + \lambda \frac{1}{N_{reg}} \sum_i p_i^* L_{reg}(t_i, t_i^*) \quad (1)$$

Here $i$ indicate "anchors" corresponding to each position of the sliding window of the input features map, $p_i$ is the probability of anchor's Objectness, and $t_i$ is the coordinates of the predicted bounding boxes. $p_i^*$ is the ground-truth label of whether an anchor is an actual object location based on Intersection over Union (IoU) with ground-truth annotations and $t_i^*$ is the coordinates of the ground-truth box corresponding to a positive anchor. The loss function is a weighted combination of a classification loss $L_{cls}$ for the binary objectness label and a regression loss $L_{reg}$ for the coordinates of the bounding box. $N_{cls}$ and $N_{reg}$ are normalization constants and λ weights the contributions of classification and regression. The classification refinement networks and the bounding box, in addition to the pooled interest regions, are trained using standard cross-entropy and regression loss functions. While Faster R-CNN has shown impressive performance on detection of everyday objects, the domain of surgical videos and surgical tools has entirely different visual characteristics. We pre-train the network on the ImageNet dataset [15], where a significant amount of data is available to learn general visual features, and then we fine tune the network on our dataset, where a smaller amount of data is labeled with the surgical tools of interest. To train the RPN, we assign a binary Objectness label to each anchor at each sliding window position of the feature map. We also assign a definite label to the anchors with an overlap greater than 0.8 with the ground-truth box or if they do not exist, an anchor or anchors with the highest Intersection over Union (IoU), and a negative label to the anchors with an IoU of less than 0.3. We adjusted the VGG-16 network to optimize model performance using the stochastic gradient descent. We modified the classification layer of the network to generate softmax probabilities over our tools. All layers are optimized for 40K iterations with a batch size of 40, and a 3×3 kernel size is used.

We perform data augmentation by randomly flipping frames horizontally. Total training time was approximately two days in an NVIDIA GPU, obtaining real-time recognition of surgical tools.

*B. Cross-Validation*

To validate our results, we use leave one out cross-validation (LOOCV) technique (shown in Fig.3). Our dataset consists of n samples of data, each time we train our model with n-1 samples and validate using 1 sample and average out the performances. In that way, we get the validation results out of all of our training data. It is stable since it reduces randomness since it tests using every sample from the original data.

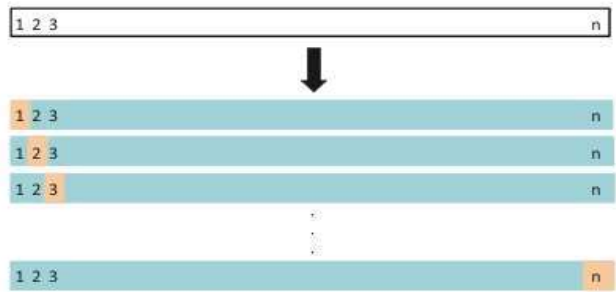

**Figure 3:** leave one out cross-validation

## IV. Experiments & Results

### A. Dataset

Our dataset consists of 16 surgery videos recorded at 25 fps in 2017 at Takatsuki Hospital, Japan. Videos were recorded, at the output of the smart glass worn by the surgeon, in MP4 format. Fig. 4 shows sample output frames captured using smart glass.

### B. Tools Annotation

We provide each frame (each of size 654x480 pixels) of the videos in JPEG format. Annotations are provided in XML templates in VOC format. Table 1 shows the number of annotated images used for CNN training. We used a 3.40 GHz CPU and NVIDIA Geforce 1080ti GPU configuration, which allowed us to run our experiments faster with less memory consumption with the cuDNN library.

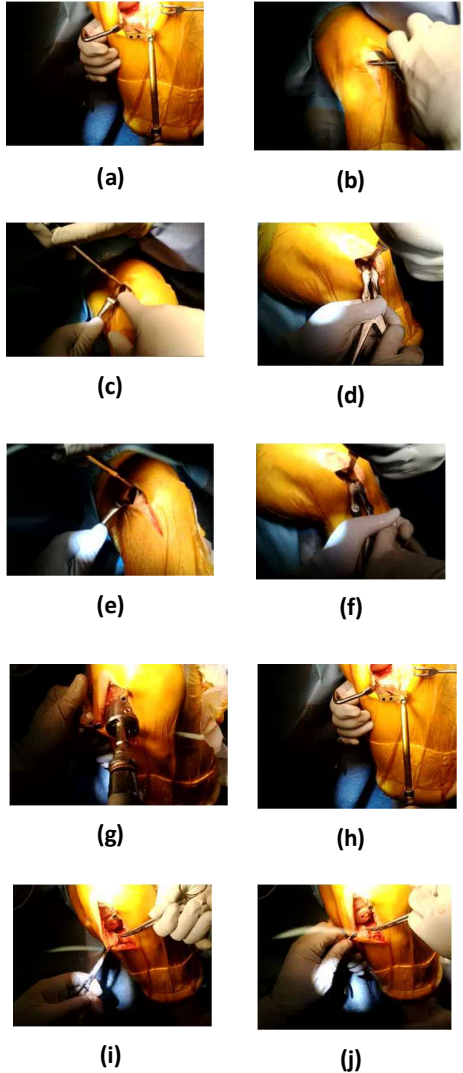

**Figure 4:** Sample output frames captured by smart glass during the surgery.

**Table I:** Number of annotated images per tool for training our CNN model

| Surgical Tools | Number of annotated images |
|---|---|
| Femoral Drill Guide | 965 |
| Spherical Mill | 1047 |
| Gap Gauge | 1116 |
| Hex Driver | 1169 |
| Tibial Template | 965 |
| Slap Hammer | 1114 |
| Posterior Resection Guide | 1057 |
| Tibial Template Medial | 935 |
| Tibial Template Nail | 1163 |
| Bearing Inserter Extractor | 1037 |
| Tibial Shim | 1193 |
| Tibial Resector Stylus | 990 |
| Tibial Impactor | 1167 |
| Tibial Groove Cutter | 1155 |
| Spigot | 1110 |
| Pin Inserter Extractor | 1164 |
| MCL Retractor | 1032 |
| IM Rod Removal Hook | 1069 |
| IM Link | 1125 |
| Femoral Impactor | 977 |
| Concise Oxford IM Awl | 926 |
| Femoral Components | 1000 |
| Chisel | 1125 |
| Cement Removal Chisel | 1049 |
| Cannulated IM rod | 1079 |
| Bone Collar Remover | 1119 |
| Anterior Bone Removal Shaft | 971 |
| Anterior Bone Mill | 929 |
| Ankle Yoke | 1070 |
| IM rod pusher | 988 |
| Tibial gap sizing spoon | 1078 |

## C. Evaluation

To test how stable the results of average precisions are, we did LOOCV for our experiment. Videos are randomly assigned to each experiment. Our results reach a mean average precision (MAP) of 88% (87.6%) shows in Table 2. It takes 9 hours to train a model with 40k iterations. We compared our experimental results with the original architecture proposed by Girschick et al. [4], Fast R-CNN using the proposals of edgeBoxes [], and the deformable parts model (DPM) suggested by Felzenszwalb et al. [5]. They are proven, state-of-the-art medical method and instrument detection in surgical videos [14]. The architecture proposed by Girschick et al. [13] scores 84% (84.48%), it takes only 4.21 hours to train a model with 40k iterations. Table 2 shows the comparison of methods where mean average precision (MAP) indicates the detection accuracy, and Detection time shows how long it takes to detect specific tool, min shows the minimum precision of the methods while max shows the maximum.

TABLE II: COMPARISON OF METHODS

| Methods | mAP | Detection time | Min | Max |
|---|---|---|---|---|
| **Proposed method** | **88 % (87.60)** | **0.075 s** | 75 | 96 |
| Fast R-CNN | 84 % (84.48) | 0.159 s | 62 | 94 |
| DPM | 76%(76.00) | 2.3 s | 57 | 84 |
| Edge+ Fast R-CNN | 20% | 0.134 | 12 | 35 |

While improvement with our approach compared to fast R-CNN seems small, but our proposed architecture has repeatedly and consistently scored more accurate scores for each series of experiments to recognize surgical tools (shown in Fig.5). Therefore, we believe that our approach has a substantial improvement over the architecture proposed by Girschick et al. [4]. However, there may be room for improvement, such as pre-training in similar flow data and the transfer of its weight to initialize our model. It could help our architecture refine its improvement over the single-mode approach. To test the effectiveness of the use of the network of regional proposals to generate proposals for the target regions, we have experimented with an alternative method of a regional proposal. Among the most popular regional proposal methods; the selective search takes about two seconds per image and generates proposals of superior quality, while Edge Box requires only 0.2 seconds per image, but compromises the quality of the proposals. With the initial experimentation, we observed that selective search did not produce higher results than the edge boxes in our data set. We have chosen to perform our experiments with edge box with the problem of time consumption. The experimental results of the use of edge box with the Fast R-CNN network reach an average precision of only 20% (table ii), while for the detection in the upper part of the 2 seconds we need 0.134 to calculate the region proposals for frames in our data set. We use a subset of our training images (each experiment configured above 2k average 2064 images) to train a model of deformable parts (DPM) (voc-release4) with memory problems. Get an average accuracy of 76%. The average detection time of the object for DPM in a test frame is 2.3 seconds.

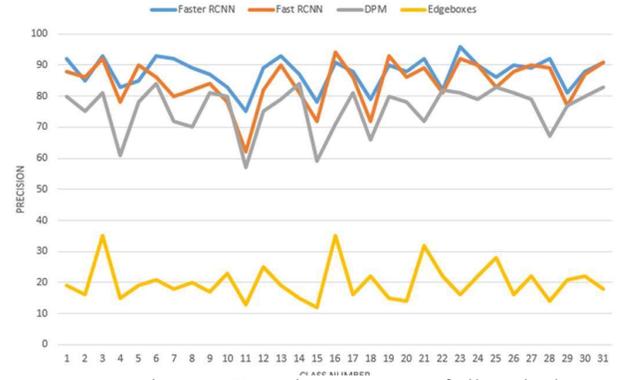

Figure 5: Detection Accuracy of all methods

In Fig. 6 Confusion matrix describes the performance of our proposed model.

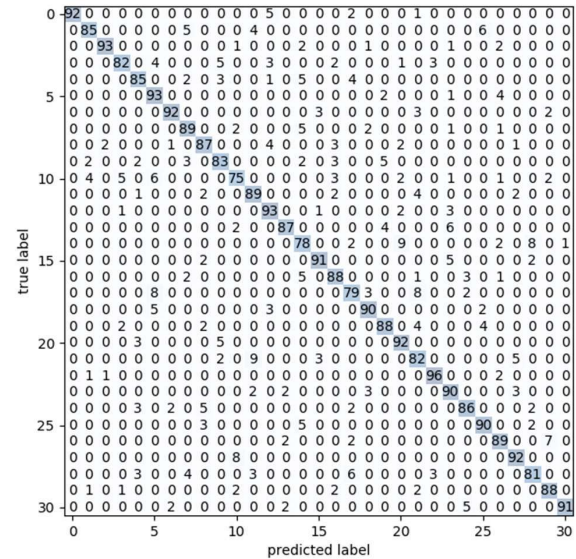

Figure 6: Confusion Matrix

## V. DISCUSSION & CONCLUSION

Our proposed method recognizes surgical tools in real-time and performs better than other studies. Understanding TKA video has not yet taken advantage of recent advances in deep neural networks. In our paper, we proposed a comprehensive approach to deep learning for real-time surgical tools recognition system for TKA. Our experimental results demonstrate that our architecture is superior to similar approaches and detection methods conventionally used in the medical industry with an average accuracy 88% per test frame calculation. Although the improvement is small, our architecture has scored more and more accurate scores for each series of experiments. With the use of YOLO [23] & SSD [17] in our model performance can be improved. We believe we can further improve our accuracy by following multimodal approach; pre-training similar flow of data and transferring the weight to initialize our model could help further refine our method. Our results demonstrate that the use of Region Proposed Network with object recognition network, whether the new multimodal proposed architecture improves accuracy and reduces calculation time for detection in each frame. We believe that our study will form a point of reference for future studies.